# SOFT LABELING STRATEGIES FOR RAPID SUB-TYPING

Grant Rosario, David Noever, and Matt Ciolino

PeopleTec, Inc., Huntsville, Alabama, USA
grant.rosario@peopletec.com

## ABSTRACT

*This challenge of labeling large example datasets for computer vision continues to limit the availability and scope of image repositories. This research provides a new method for automated data collection, curation, labeling, and iterative training with minimal human intervention for the case of overhead satellite imagery and object detection. The new operational scale effectively scanned an entire city (68 square miles) in grid search and yielded a prediction of car color from space observations. A partially trained yolov5 model served as an initial inference seed to output further, more refined model predictions in iterative cycles. Soft labeling here refers to accepting label noise as a potentially valuable augmentation to reduce overfitting and enhance generalized predictions to previously unseen test data. The approach takes advantage of three unique real-world instances, each of which demonstrate how pixel values alone can provide enough sub-type information of a cropped image to determine if a car is white or colorful, if a building's roof is blue, or thirdly, if a crude oil tank is full or empty, thus completing an end-to-end pipeline without overdependence on human labor.*

## 1. INTRODUCTION

One advance for machine learning research and computer vision problems would automate the laborious generation of expert labels for object types. The traditional labeling approach divides the data among multiple labelers and pools their bounding boxes, confidences, and object classes. An approximate rule for building robust detectors involves at least 10,000 examples when training from scratch [1]. Extending the features of a pre-trained model (e.g., transfer learning) can reduce the target count 10-fold, but even this labeling workload remains labor-intensive for 1000 examples. One suggested crowd-sourced price for object labels ranges between $0.03-0.04 per label or $40-$400 per class [2]. For the 60 categories or object types included in a large overhead image dataset like xView [3], the cost of its million object pre-processing might exceed $100,000 using current cloud technology and three workers per image. Previous research [4-11] has focused on automating and strategically lightening the human contributions to address this tedious labor challenge.

The current work explores a concrete example of the semi-supervised method [4-6]. The application demonstrates using soft labels with noise. Akin to bootstrapping from a partially trained vision model, the technique records the bounding box locations for each training image, then retrains repeatedly, thus accepting the output (including false positives and negatives) as inputs for the next cycle. While this semi-supervision strategy still needs an initial human labeler (to seed the first 100 examples), repeated and expanded training processes automate the refinement of model performance (e.g., F1, mean average precision, etc.). The strategy abandons some initial curation approaches, particularly those that identify low-confidence class assignments or continuously prune dataset sizes to include better labels. The noise in this example bolsters the final model's ability to generalize, rather than memorize, the training data. Similar to augmentation noise, the scale of larger datasets may contribute to model improvement. In survey literature [12-15], researchers have embraced alternatives to assemble large datasets, including synthetic placement of objects in pre-determined boxes, Zhou et al. [10] called the benefits of this pre-training approach not as much of a "free lunch" but more of an efficiency-driven "cheaper lunch." We embrace the "cheaper but not free lunch" analogy in demonstrating a semi-supervised approach to generating plausible label categories from partially trained model outputs.

We explore soft labels with semi-supervised learning and label noise for the specific task to sub-type a broader single class. Like face detection, the classic sub-typing problem assumes a detector can find a face but must now behave more specifically to identify the person or sub-type. In our case, we present three novel examples of identifying object sub-types from overhead satellite photos: white cars, crude oil tank capacity, and blue roofs. Unlike the facial recognition problem, our model freezes almost all standard and distinguishing features of the overhead object detector: texture, resolution, shape, size, and view angle. Instead, we specialize in binary labeling and assignment ("white" vs. "non-white" colors, etc.). While we conducted this research over three scenarios, for the sake of simplicity the bulk of this paper will focus on the "white" vs "non-white" car problem while still providing results and minor data discussion of the other two scenarios. The model framework and experiment method stayed consistent across each of the scenarios.

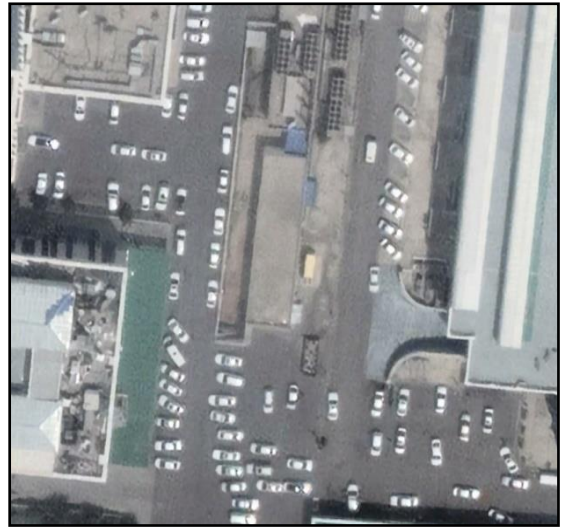

Figure 1. Overhead image of Turkmenistan's capital and enforced white car rule. Ashgabat location, 37.914789°,58.383500°

In contrast to the hierarchical feature extraction typical of deep convolutional neural networks, the features of shape and texture become subordinate to the network weights governing color only. A beneficial outcome of this dataset further propagates the color assignment in a deterministic way from the pixel map. We demonstrate taking a cropped version of all vehicles, then assigning "white" vs. "non-white" using a simple pixel grade with white sampled in grayscale as near 255. A similar method is done for detection of blue building roofs in order to only assign "blue" labels by taking a cropped version of buildings and assigning "blue" when it falls in the relevant HSV color range. Regarding crude oil tank volume, we similarly analyze a cropped version of a tank, but rather than focus on color, we label a tank as "full" when the mass of its exterior shadow is at least 30% larger than its interior shadow, otherwise it is labeled "empty" (Figure 2).

The white car dataset credits an unusual 2018 law from Turkmenistan's President Gurbanguly Berdymukhammedov. The leader of the former Soviet republic banned all non-white cars from its capital city, Ashgabat [16]. With white considered "lucky," this mandate follows previous regulations that outlawed tinted windows (2014), personalized license plates (2014), and restricted imported engine sizes (2015). Non-white cars are subject to seizure on the streets, and owners must agree to repaint existing vehicles at an individual cost of the average annual salary. Given the low initial distribution of white cars, we collect training data in a roughly similar urban environment, Tajikistan's capital, Dushanbe. The methods section outlines the data collection, curation, and labeling strategy in detail, along with the Yolo-family model [1] and reporting metrics for overall performance. A similar reasoning can be given for the prevalence of blue roofs on the Korean peninsula; however, this is not due to any law but rather more likely just a cultural view of affluence since the presidential residence has a blue roof [19].

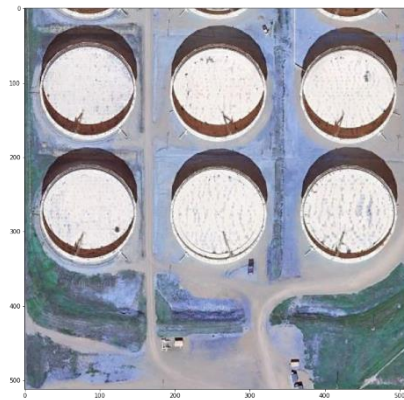

Figure 2. Overhead image of crude oil tanks with shadows clearly visible as useful indicators of current occupied volume.

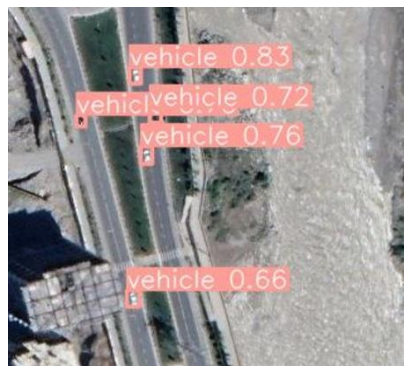

Figure 3. Example labeling for first pass without distinguishing sub-types (e.g., all vehicles shown with confidence)

## 2. METHODS

The machine learning pipeline collects and labels overhead imagery using automatic Google Earth navigation. The method covers city-scale image collection by touring automatically through a grid (68 sq. km) using vertically and horizontally overlapping grids. The approximate image altitude is 0.6 km, so a sedan vehicle comprises roughly 100 (20x5) pixels for identification. Depending on the view's location and date, the imagery traces to WorldView-03 satellites in color, with an approximate ground resolution of 0.3 m per pixel (ground sample distance, or GSD). Most imagery dates to 2021 depending on the location, satellite availability, and clear, non-cloudy weather. Using the National Imagery Interpretability Rating Scale (NIIRS), the imagery can be scored based on its resolution and suitability to fulfill typical sub-type analysis. Identifying car shapes (sedan, SUV, etc.) typically requires at least 1 m GSD (NIIRS=4-5). Identifying windshields requires 0.5 m (NIIRS=6), and resolving car mirrors needs at least 0.15 m aerial resolution (NIIRS=8).

### 2.1. Datasets

We assembled an overhead (nadir) image collection using a desktop Google Earth application that follows a grid search across a designated city area. We limit each capture to 416x416 imagery to optimize the labeling speed expected in yolov5 implementations. The initial pass collected an urban region of interest in Dushanbe, Tajikistan. The white car dataset included 6570 cars in 477 images, or an approximate average of 14 vehicles per scene, significantly varying from frame to frame (Figure 3). We derive the test set from Turkmenistan's capital, Ashgabat after authorities enforced the all-white car rule (2018). We use the history capability in Google Earth to date the images collected and document the changes over time in semi-automated labeling cycles. The blue roof dataset was gathered similarly but from the Korean peninsula and contains ~80000 roofs in 367 images while the oil tank dataset includes 3772 tanks in 1994 images.

### 2.2. Model Framework

We attached soft labels to the automated collection of overhead imagery using a pre-trained vehicle detector based mainly on a heavily augmented xView dataset [3] and a yolov5 model [1]. This custom model was trained originally on three classes for moving objects

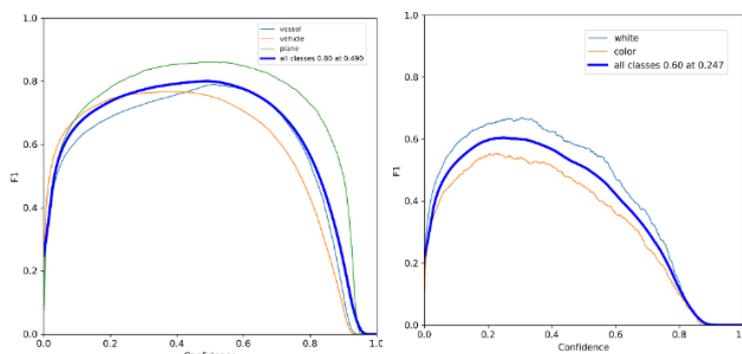

Figure 4. F1 metrics for xView augmented vehicle model peaks at 0.2 confidence level near 0.74 in the bootstrap starter model (vehicle, ship, plane) and peaks at 0.6 for distinguishing white vs. color cars in the soft-label sub-type extension

(vehicles, planes, and ships) extracted from pre-trained models and 34,252 images collected from parking lots, marine ports, and airports. Approximately one-third of the pictures were background, meaning they contained no examples but contributed to the training "null" set. The number of labeled vehicles in this composite dataset totaled 174,779 vehicles in 22,458 images (or around eight vehicles per image with a broad variance between frames). The maximum number of cars labeled in a single image totaled 720.

This bootstrapped starting model subsequently accelerates the tedious labeling process. Undoubtedly a tradeoff exists between ever more comprehensive data and labeling noise. Even incorrect labels may increase the model's generalizability. Subsequent label noise, or soft tags, may reduce the model's bias, overfitting, or memorization of training examples. Previous workers have proposed and verified the practical benefits of this approach in other contexts, such as language modeling. The trained PyTorch weights provide inference predictions on unseen test data, from which newly labeled samples can provide model feedback in subsequent cycles with increasingly more extensive datasets. This virtuous cycle necessarily admits in mis-labeled samples that a human might filter out from future training data.

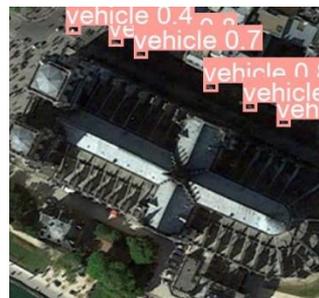

Figure 5. Example validation output from the pre-trained model

### 2.3. Metrics

In unbalanced datasets typical of overhead satellite imagery [3], the F1 score penalizes false positives and negatives as shown left in Figure 4. The xView dataset [3] labels almost half of the total objects using just two tags (building and cars) out of the 60 class ontology. We augment the moving objects only (planes, ships, and vehicles) with a balanced starting dataset by including other custom datasets and culling over-sampled results. Yolov5 [1] offers a built-in feature to subset any group of classes and suppress minority classes when performing inference. For a balanced dataset, the mean average precision (mAP @ 0.5 and @0.95) also flows directly from the internal metrics reported by yolov5. Precision refers to the correct location of the anchor corners for bounding boxes, which relies on repeatable labelers who subjectively encapsulate the object without excessive or deficient surrounding pixels. We show the training and validation metric panels for comparison before and after soft labels.

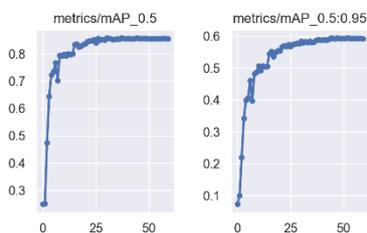

Figure 6. Mean average precision (mAP) for 3-class bootstrap from xView moving object data

## 3. RESULTS AND DISCUSSION

Before applying a bootstrapped model, we verified the accuracy of a general vehicle detector. Figure 6 summarizes the mAP > 0.8 for moving object detection in planes, ships, and vehicles. All the yolov5 models showed plateauing metrics after 50 epochs so training terminated. Missing cars or confusing labels with background images proved to be the most consistent error for the vehicle class. The output of the bootstrap model then generated the next cycle of new training data by looping over another "test-to-train" iteration. It is worth noting that a modern GPU (Nvidia RTX 5000) can process 10-50 frames per second using the yolov5 small model (optimized weights around 14 Mb total file size) [1]. When comparing these mAP values to those found from the original 60 classes in xView, the previously best values reported for that more complex detection and multi-class challenges fell below 0.2.

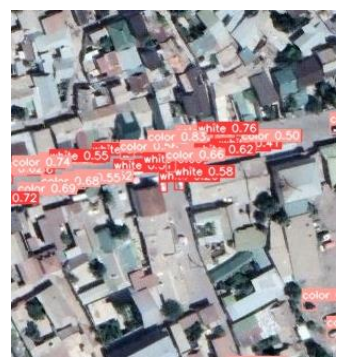

Figure 7. Sub-typing for white vs. non-white vehicles

After running an inference pass on the Dushanbe city, the bootstrap model automatically scanned, identified, and labeled 6570 vehicles in 473 images of the Tajikistan capital. Since yolov5 can crop the vehicles into thumbnails while also creating bounding boxes, a quick scan of the extracted images showed the "white" cars as a separate class. Pixel values saturated around 255 in grayscale identify the lightest colored cars. The resulting inference output in one category ("vehicles") thus provides a simple bifurcation means, with pixel values as the automated guide. Given the two new classes

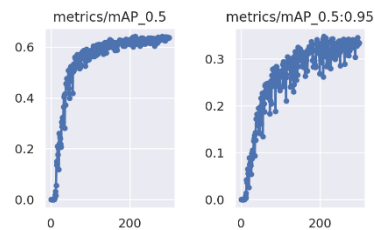

Figure 8. Mean average precision (mAP) for 2-class subtype model from soft label learning

("white" vs. "color"), a new training cycle compounds the previous results for vehicle identification with the further sub-types based only on color (Figure 7). The example precisions shown in Figure 7 labels and the labeled validation image (Figure 8) along a road the mixture of white and non-white cars expected in a regionally specific capital city (Dushanbe) that does not enforce car color as a regulated requirement. Figure 4 shows the F1 score peaks at 0.6 at a confidence interval of 0.247. Figure 8 verifies a visual capability to learn vehicle types, then automate the previously tedious task of sub-typing and labeling a large urban scene with minimal human intervention in data collection, curation, labeling, and retraining.

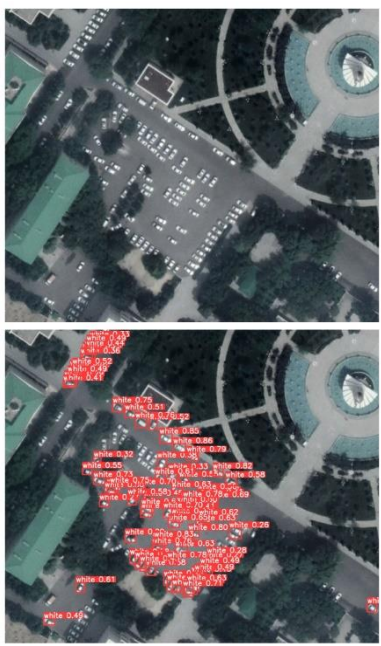

Figure 9. The soft label model (lower) identifies 75 white cars, no color cars vs. a human labeler who identifies 106 white cars (72%).

When the soft label model predicted car color using test data in Ashgabat, Turkmenistan (latitude=37.941442°, longitude=58.385896°), the model found only white cars as expected and identified no colors (Figure 9). While 75 cars (out of 106) were correctly put in bounding boxes with labels, the yolov5 and other anchor-dependent models without overlapping boxes tend to undercount crowded scenes.

To verify the two-label model, we applied the trained soft labels to an April 2018 image in Ashgabat before the broad enforcement of white cars only. Figure 10 shows a parking lot (latitude=37.940787°, longitude=58.380398°) with around 10% of the vehicle population in the category of non-white cars.

The metrics and results from the same soft-labeling strategy for assessing oil tank volume capacity and blue roof recognition are shown Figure 11.

## 4. CONCLUSIONS

Soft labels that accept noisy inputs and a partially trained model represent a scalable approach to assigning iterative labels with minimal human intervention. Using a human-labeled model like xView and filtering to a particular single class first proved helpful in seeding the initial automatic labels. The problem selection to remove all features except color from a real-world dataset provides a convenient test of both feature importance and learning for other sub-types. In effect, color changes alone signal the class without proper weights describing shape, size, or vehicle style in this case. While the F1 score for the single category (vehicles) was high (0.8), the sub-type

(white vs. color) achieved reasonable values (0.6) and greatly exceeded the highest multi-class value (0.2) reported in the original xView challenge for overhead vehicle identification. The original contributions of this research highlight the future potential for automated data collection, curation, labeling, and iterative training with minimal human intervention. The new operational scale effectively scanned an entire city (68 square miles) in grid search and yielded a prediction of car color from space observations.

Banning all colored cars from a complex urban environment presents an intriguing use case for examining automated labeling strategies. The method embraces the speed of labeling 10,000 or more images in an hour without incurring the cost of hiring multiple human labelers to engage in tedious or error-prone steps at little or no incremental value. If one accepts that training a model from scratch requires 10,000 examples per class (or, if transfer learning, 1000 examples per class), the current approach extends a single class rapidly to use heuristics like pixel values for finishing the sub-typing task. We anticipate this example generalizes to other real-world datasets of practical importance in global resource monitoring.

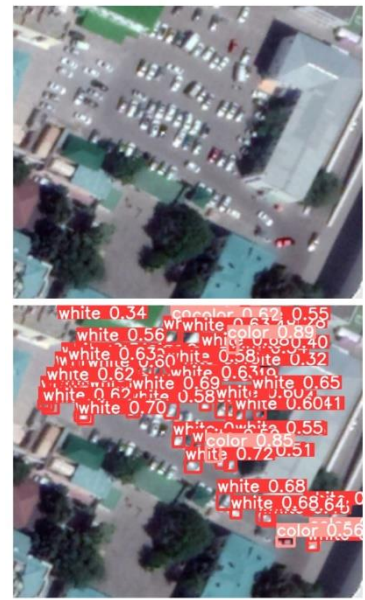

Figure 10. The soft-label model (lower) identified 7 colored cars in a background of 62 white cars, April 2018, prior to the legal restrictions.

One immediate candidate for soft-labels and subtypes includes the routine and important need to inventory large POL facilities (Petroleum, Oil, and Lubricants Storage) [17]. Since many of these storage tanks have floating tops, satellite observations and appropriate algorithms have combined to render useful monitoring of global reserves near large ports. In a soft-label example, two-classes might represent "full" and "partial" volumes and similarly take advantage of the tank detector itself as the first bootstrap for refining the search. A second useful example might include the global interest in finding and counting solar panels. Stanford's Deep Solar project [18] presented the goal to count every US solar panel and identified in 2019 a total of 1.47 million solar installations, a number much higher than traditional methods like surveys or manufacturing data. Once a general model for "solar panel" gets trained, the sub-typing to residential or commercial solar installations becomes a candidate for soft-labels and sub-type methods.

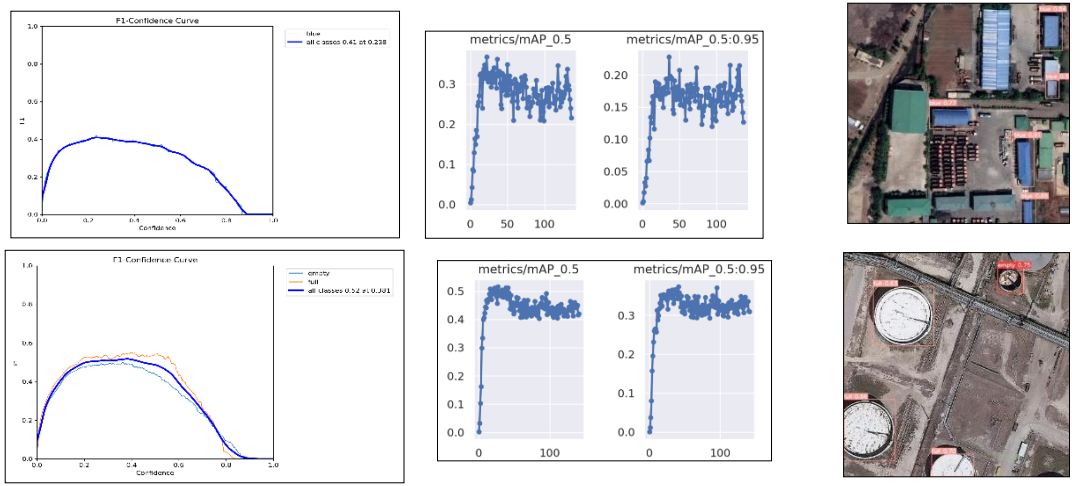

Figure 11. Top Left: F1 score peaks at 0.4 for blue roof detection. Top Middle: Mean-avg precision for single-class blue roof subtype. Top Right: Blue roof subtype result Bottom Left: F1 score peaks at ~0.5 for empty vs full oil tank detection. Bottom Middle: Mean-avg precision for 2-class oil tank subtype. Bottom Right: "Full" vs "empty" subtype result.

## ACKNOWLEDGMENTS

The authors would like to thank the PeopleTec Technical Fellows program for its encouragement and project assistance.

## REFERENCES

[1] Jocher, G., Stoken, A., Borovec, J., Changyu, L., Hogan, A., Diaconu, L., ... & Yu, L. (2020). ultralytics/yolov5: v3. 1-bug fixes and performance improvements. Version v3, 1.

[2] Amazon SageMaker Data Labeling (2022), https://aws.amazon.com/sagemaker/data-labeling/pricing/

[3] Lam, D., Kuzma, R., McGee, K., Dooley, S., Laielli, M., Klaric, M., ... & McCord, B. (2018). xview: Objects in context in overhead imagery. *arXiv preprint arXiv:1802.07856*.

[4] Wu, Z., Bodla, N., Singh, B., Najibi, M., Chellappa, R., & Davis, L. S. (2018). Soft sampling for robust object detection. arXiv preprint arXiv:1806.06986.

[5] Xu, M., Zhang, Z., Hu, H., Wang, J., Wang, L., Wei, F., ... & Liu, Z. (2021). End-to-end semi-supervised object detection with soft teacher. In *Proceedings of the IEEE/CVF International Conference on Computer Vision* (pp. 3060-3069).

[6] Li, J., Xiong, C., Socher, R., & Hoi, S. (2020). Towards noise-resistant object detection with noisy annotations. *arXiv preprint arXiv:2003.01285*.

[7] Tang, Y., Chen, W., Luo, Y., & Zhang, Y. (2021). Humble teachers teach better students for semi-supervised object detection. In *Proceedings of the IEEE/CVF Conference on Computer Vision and Pattern Recognition* (pp. 3132-3141).

[7] Peng, J., Bu, X., Sun, M., Zhang, Z., Tan, T., & Yan, J. (2020). Large-scale object detection in the wild from imbalanced multi-labels. In *Proceedings of the IEEE/CVF conference on computer vision and pattern recognition* (pp. 9709-9718).

[8] Yu, Y., Yang, X., Li, J., & Gao, X. (2022). Object Detection for Aerial Images with Feature Enhancement and Soft Label Assignment. *IEEE Transactions on Geoscience and Remote Sensing*.

[9] Zhang, C. B., Jiang, P. T., Hou, Q., Wei, Y., Han, Q., Li, Z., & Cheng, M. M. (2021). Delving deep into label smoothing. *IEEE Transactions on Image Processing*, *30*, 5984-5996.

[10] Zhou, D., Zhou, X., Zhang, H., Yi, S., & Ouyang, W. (2020, August). Cheaper pre-training lunch: An efficient paradigm for object detection. In *European Conference on Computer Vision* (pp. 258-274). Springer, Cham.

[11] Du, Y., Chen, Z., Jia, C., Li, X., & Jiang, Y. G. (2021, August). Bag of Tricks for Building an Accurate and Slim Object Detector for Embedded Applications. In *Proceedings of the 2021 International Conference on Multimedia Retrieval* (pp. 519-525).

[12] Mao, H. H. (2020). A survey on self-supervised pre-training for sequential transfer learning in neural networks. *arXiv preprint arXiv:2007.00800*.

[13] Cai, L., Zhang, Z., Zhu, Y., Zhang, L., Li, M., & Xue, X. (2022). BigDetection: A Large-scale Benchmark for Improved Object Detector Pre-training. In *Proceedings of the IEEE/CVF Conference on Computer Vision and Pattern Recognition* (pp. 4777-4787).

[14] Liu, L., Ouyang, W., Wang, X., Fieguth, P., Chen, J., Liu, X., & Pietikäinen, M. (2020). Deep learning for generic object detection: A survey. *International journal of computer vision*, *128*(2), 261-318.

[15] Zoph, B., Ghiasi, G., Lin, T. Y., Cui, Y., Liu, H., Cubuk, E. D., & Le, Q. (2020). Rethinking pre-training and self-training. *Advances in neural information processing systems*, *33*, 3833-3845.

[16] Sergeev, A., (2018), Turkmenistan President Bans All Non-White Cars From Capital, https://www.motor1.com/news/226932/turkmenistan-president-bans-black-cars/

[17] Dean, J.C., Geusic, S., and Regin, T.M. (2021) Petroleum, Oil, And Lubricants (POL) Storage And Distribution Systems Knowledge Area https://www.wbdg.org/ffc/dod/cpc-source/petroleum-oil-lubricants-storage-distribution-systems-knowledge-area

[18] Brown, M. (2019) How Do You Count Every Solar Panel in the U.S.? Machine Learning and a Billion Satellite Images, Engineering.com, https://www.engineering.com/story/how-do-you-count-every-solar-panel-in-the-us-machine-learning-and-a-billion-satellite-images

[19] Woo, John (2022) In Korea, the Blue House Era Ends. https://kpcnotebook.scholastic.com/post/korea-blue-house-era-ends

**Authors**

| | |
|---|---|
| **Grant Rosario** has research experience in embedded applications and autonomous driving applications. He received his Masters from Florida Atlantic University in Computer Science and his Bachelors from Florida Gulf Coast University in Psychology. | 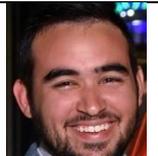 |
| **David Noever** has research experience with NASA and the Department of Defense in machine learning and data mining. He received his BS from Princeton University and his Ph.D. from Oxford University, as a Rhodes Scholar, in theoretical physics. | 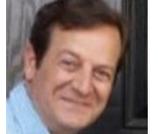 |
| **Matt Ciolino** has research experience in deep learning and computer vision with specializations in super-resolution, text augmentation, and training data curation methods. He currently pursues his Masters from Georgia Institute of Technology and received his Bachelor's from Lehigh University in Mechanical Engineering. | 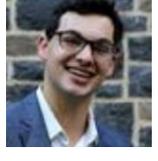 |